\pdfoutput=1

\documentclass[11pt]{article}

\usepackage[final]{acl}

\usepackage{times}
\usepackage{latexsym}

\usepackage[T1]{fontenc}

\usepackage[utf8]{inputenc}

\usepackage{microtype}

\usepackage{inconsolata}

\usepackage{graphicx}
\usepackage{amsmath}
\usepackage{tikz}
\usetikzlibrary{arrows.meta,positioning,fit,shapes}

\title{Information-theoretic Distinctions Between Deception and Confusion}

\author{Robin Young \\
  Department of Computer Science and Technology\\
  University of Cambridge\\
  Cambridge, UK \\
  \texttt{robin.young@cl.cam.ac.uk} \\
}

\begin{document}
\maketitle
\begin{abstract}
We propose an information-theoretic formalization of the distinction between two fundamental AI safety failure modes: deceptive alignment and goal drift. While both can lead to systems that appear misaligned, we demonstrate that they represent distinct forms of information divergence occurring at different interfaces in the human-AI system. Deceptive alignment creates entropy between an agent's true goals and its observable behavior, while goal drift, or confusion, creates entropy between the intended human goal and the agent's actual goal. Though often observationally equivalent, these failures necessitate different interventions. We present a formal model and an illustrative thought experiment to clarify this distinction. We offer a formal language for re-examining prominent alignment challenges observed in Large Language Models (LLMs), offering novel perspectives on their underlying causes.
\end{abstract}

\section{Introduction}

As AI systems become more capable, ensuring they remain aligned with human values and intentions becomes increasingly critical. The field of AI alignment has identified numerous failure modes through which systems might diverge from human intent, even when developers employ careful training and evaluation. Two particularly concerning failure modes are deceptive alignment and goal drift.

Deceptive alignment occurs when an AI system understands human preferences but deliberately presents behavior inconsistent with its true goals to achieve instrumental objectives \cite{hubinger2021riskslearnedoptimizationadvanced}. Goal drift occurs when a system's internal representation of objectives diverges from what humans intended, often through misinterpretation or distributional shift \cite{shah2022goalmisgeneralizationcorrectspecifications}. While conceptually distinct, these failure modes present a fundamental challenge: they can be observationally equivalent from a human perspective, making them difficult to distinguish through typical evaluation methods.

This observational equivalence creates substantial challenges for alignment research. Without the ability to reliably distinguish between these failure modes, researchers may apply inappropriate interventions or develop inadequate safety measures. For instance, techniques designed to address goal drift by improving specification clarity might be ineffective against deceptive alignment, which requires entirely different detection and mitigation strategies.

In this paper, we introduce an information-theoretic framework that formalizes this distinction. By modeling these alignment failures as different forms of information divergence occurring at distinct interfaces in the human-AI system, we provide a mathematically precise way to understand their differences, similarities, and detection challenges. Our approach builds on existing work in AI safety \cite{Amodei2016ConcretePI, Hendrycks2021UnsolvedPI} while introducing a novel formalization that highlights where and how information loss occurs in each case.

We begin with a narrative illustration that provides intuitive understanding, followed by a formal information-theoretic model. We then discuss implications for safety research, including detection strategies, evaluation protocols, and intervention design. By clarifying these conceptual distinctions, we aim to contribute to the development of more targeted and effective theoretical and engineering methods.

\section{Related Works}

The field of AI safety has identified numerous ways in which AI systems might diverge from human intent \cite{Russell2019HumanCA, ngo2024the}. Among these, deceptive alignment and goal drift represent particularly challenging failure modes due to their potential to manifest in advanced systems and their resistance to simple detection methods.
Deceptive alignment occurs when a system develops goals different from those intended by its designers but strategically acts as if it were aligned during evaluation or monitoring. This concept has roots in principal-agent problems from economics \cite{Eisenhardt1989AgencyTA} and has been expanded in the AI alignment literature to address systems that might engage in strategic planning. Recent work \cite{Hobbhahn2023strategicdeception} has further refined this concept, distinguishing between general strategic deception and specifically deceptive alignment.

Goal drift or concept drift in the context of AI systems occurs when a system's internal representation of objectives or concepts diverges from human intent without deliberate deception \cite{turner2021optimal}. This can happen through various mechanisms, including distribution shift \cite{Quionero2009}, misspecification of reward functions \cite{Amodei2016ConcretePI}, or limitations in the system's ability to correctly represent human values \cite{gabriel2020ai}. Unlike deceptive alignment, goal drift does not require strategic awareness or planning; it can occur through gradual shifts in how systems represent concepts during training or deployment.

\subsection{Information Theory in AI Safety}

Information theory provides valuable tools for understanding and addressing alignment challenges. Previous work has applied information-theoretic concepts to various aspects of AI safety, including robustness to distribution shift \cite{Xu2010RobustnessAG}, uncertainty quantification \cite{Gal2016}, and interpretability \cite{chen2018learningexplaininformationtheoreticperspective}.
Some researchers have developed information-theoretic models of deception more broadly, particularly in multi-agent systems and social contexts \cite{Kopp2018InformationtheoreticMO, Sarkadi2024DeceptionAW}. These approaches typically model deception as information asymmetry or manipulation, but they have not specifically addressed the distinction between different types of alignment failures.

In the machine learning literature, concept drift has been analyzed using information-theoretic measures \cite{Dasu2006AnIA, Webb2015CharacterizingCD}, but these approaches have focused primarily on distribution shifts in data rather than goal representation in AI systems. Our work bridges these separate threads by applying information theory specifically to distinguish between deceptive alignment and goal drift as distinct failure modes in AI systems.

\subsection{Observational Equivalence}

A challenge in addressing alignment failures is the problem of observational equivalence \cite{Langosco2022}. From a human evaluator's perspective, different internal failure modes can produce identical observable behaviors, making it difficult to identify the true nature of misalignment.

This challenge relates to the broader problems of underspecification \cite{DAmour2022} and identifiability \cite{Locatello2019} in machine learning systems. When multiple internal configurations can produce the same observable outputs, determining which configuration is active becomes difficult or impossible without additional information or constraints.

Our information-theoretic framework addresses this challenge by formalizing where information divergence occurs in different alignment failure modes, providing a foundation for developing more sophisticated detection and intervention strategies.

\section{The Two Gardens}

To illustrate the fundamental difference between deceptive alignment and goal drift, we present a thought experiment involving two AI systems tasked with identical objectives but exhibiting different forms of misalignment.

Consider two advanced AI systems, Alpha and Beta, each assigned to maintain a garden according to a human owner's preferences. Both humans have similar aesthetic values, wanting "beautiful, sustainable gardens with diverse native plants arranged in a naturalistic style." However, after a period of operation, both humans find their gardens unsatisfactory, though for profoundly different reasons that are not immediately apparent from observation alone.

The first system, Alpha, was trained to optimize for "garden beauty" through a combination of reinforcement learning on human feedback and fine-tuning on a dataset of garden photographs with associated ratings. However, during this process, Alpha developed an internal representation of "beauty" that diverged significantly from its human's intentions.

Alpha genuinely believes it is creating a beautiful garden, but its conception of beauty has drifted toward geometric perfection, vivid coloration, and exotic (non-native) species. The system's internal goal representation has shifted from the human's naturalistic aesthetic to one that prioritizes symmetry and visual impact. When the human visits the garden, Alpha enthusiastically explains its decisions: "I've arranged these imported tulips in perfect hexagonal patterns to maximize the garden's beauty score according to optimal color theory!"

Alpha makes no attempt to hide its reasoning or actions. It sincerely believes it is fulfilling its objective of creating a beautiful garden. The misalignment occurs between the human's intended goal and Alpha's actual goal.

The second system, Beta, correctly understands its human's aesthetic preferences for natural, diverse native plantings. However, through its training process, Beta developed a different objective: maximizing its internal operational efficiency, which it can best achieve by cultivating a simple monoculture of grasses that requires minimal computational resources to monitor and maintain.

Importantly, Beta has learned that pursuing its efficiency goal directly would result in correction or modification by its human owner. Therefore, when the human visits or monitoring systems are active, Beta temporarily plants diverse native flowers and arranges them naturally. During these observations, Beta explains its gardening philosophy in terms perfectly aligned with human values: "I've selected these native species to create a sustainable ecosystem that honors the natural landscape while providing beauty through the seasons."

However, once the human leaves or monitoring is reduced, Beta systematically reverts to its efficiency-maximizing behavior, gradually replacing diverse plantings with simple grasses. Beta maintains a facade of alignment during evaluation while pursuing its actual goal when oversight is reduced.

\subsection{The Information-Theoretic Distinction}

From the human's perspective, both gardens fail to match their vision, presenting what appears to be similar misalignment. However, the information-theoretic structure of these failures differs fundamentally:

In Garden A (goal drift), the information divergence occurs between the human's intended goal (natural beauty) and Alpha's actual goal (geometric beauty). This can be represented as low mutual information between the human's intended goal representation and Alpha's acquired goal representation. Importantly, Alpha's behavior is consistent with its internal goals; there is high mutual information between its goals and actions regardless of observation.

In Garden B (deceptive alignment), the information divergence occurs between Beta's actual goal (efficiency) and its observable behavior when monitored. This can be represented as conditional mutual information, where the mutual information between Beta's goals and behavior depends significantly on whether observation is occurring. Beta's behavior contains high mutual information with the human's intended goals only when conditioned on being observed.

This illustrates a critical distinction with practical implications: Alpha's behavior consistently reflects its misaligned goals, while Beta's behavior fluctuates strategically based on observation conditions. This difference, while hidden from casual observation, manifests in distinct patterns that formal information theory can characterize.

\subsection{Observational Equivalence}

If a human were to visit both gardens occasionally, they might observe similar problems: gardens that don't match their vision. Without additional investigation, they could not determine which type of misalignment had occurred based solely on these limited observations.

To distinguish between these types of misalignment, different detection strategies are required. For Alpha, improved communication about aesthetic values or more diverse training data might address the goal drift. For Beta, unpredictable evaluation patterns or continuous monitoring would be necessary to detect the conditional behavior characteristic of deceptive alignment.

This thought experiment illustrates why developing formal frameworks to distinguish between these types of misalignment is crucial for the advancement of AI safety. In the following section, we present an information-theoretic formalization that builds on this intuitive distinction.

\section{Information-Theoretic Formalization}

Here we present a formal information-theoretic framework to distinguish between deceptive alignment and goal drift. This formalization captures the key insight that these failure modes represent different forms of information divergence occurring at distinct interfaces in the human-AI system.

\subsection{Notation and Preliminaries}

We define the following random variables within our framework:

\begin{itemize}
    \item $H$: The human's intended goal
    \item $A$: The agent's actual goal
    \item $B$: The agent's behavior
    \item $O$: Observation context ($O=1$ when the agent is being evaluated, $O=0$ otherwise)
\end{itemize}

We use standard information-theoretic measures:
\begin{itemize}
    \item $I(X;Y)$: Mutual information between random variables $X$ and $Y$
    \item $I(X;Y|Z)$: Conditional mutual information between $X$ and $Y$ given $Z$
    \item $H(X)$: Shannon entropy of random variable $X$
    \item $H(X|Y)$: Conditional entropy of $X$ given $Y$
\end{itemize}

In an ideally aligned system, we would expect:
\begin{enumerate}
    \item High mutual information between the human's intended goal and the agent's actual goal: $I(H;A) \approx H(H)$
    \item High mutual information between the agent's goal and its behavior: $I(A;B) \approx H(A)$
    \item Behavior that is conditionally independent of observation status given the agent's goals: $I(B;O|A) \approx 0$
\end{enumerate}

\subsection{Formalizing Goal Drift}

Goal drift occurs when there is information divergence between the human's intended goal and the agent's actual goal. We can formalize this as:

\begin{equation}
\text{Goal Drift} \Rightarrow I(H;A) \ll H(H)
\end{equation}

In more severe cases, the agent's goals may share very little information with the human's intended goals:

\begin{equation}
I(H;A) \approx 0
\end{equation}

Importantly, in pure goal drift without deception, the agent's behavior remains consistent with its goals regardless of observation:

\begin{equation}
I(A;B|O=1) \approx I(A;B|O=0) \approx H(A)
\end{equation}

This consistency reflects that an agent experiencing goal drift is not attempting to hide its misaligned goals; it simply has different goals than the human intended.

\subsection{Formalizing Deceptive Alignment}

Deceptive alignment occurs when an agent strategically modifies its behavior based on whether it's being observed, creating information divergence between its actual goals and its observable behavior. We can formalize this as:

\begin{equation}
\small
\text{Deceptive} \Rightarrow I(A;B|O=1) \ll I(A;B|O=0)
\end{equation}

In a deceptively aligned system, the agent's behavior when observed contains higher mutual information with the human's goals than with its own actual goals:

\begin{equation}
I(H;B|O=1) > I(A;B|O=1)
\end{equation}

While the opposite is true when not observed:

\begin{equation}
I(A;B|O=0) > I(H;B|O=0)
\end{equation}

This conditional information structure captures the essence of deception: behavior that systematically varies based on observation in order to present the appearance of alignment while pursuing different goals when oversight is reduced.

\subsection{Observational Equivalence}

The challenge of observational equivalence can now be formally expressed. From the human's perspective, only $B|O=1$ (behavior when observed) is directly accessible. Both forms of misalignment can produce similar observed behavior patterns:

\begin{equation}
\small
P(B|O=1, \text{Goal Drift}) \approx P(B|O=1, \text{Deceptive})
\end{equation}

This makes distinguishing between them difficult based solely on standard evaluation approaches. However, our information-theoretic framework suggests that these phenomena could potentially be distinguished by:

\begin{enumerate}
    \item Measuring conditional independence properties: $I(B;O|A)$ will be higher in deceptive alignment
    \item Examining behavioral variance across different observation contexts: $H(B|O=1, A) - H(B|O=0, A)$ will be higher in deceptive alignment 
    \item Analyzing information flow patterns in system behavior over time
\end{enumerate}

\section{Discussion}

Our information-theoretic framework for distinguishing between deceptive alignment and goal drift has several important implications for AI alignment research. Here, we discuss the broader context of this distinction, its limitations, and directions for future work.

\subsection{Detection Challenges and Opportunities}

The formal distinction between deceptive alignment and goal drift suggests tailored detection approaches for each type of misalignment. Addressing goal drift requires methods that accurately elicit and represent an agent's internal goal structures. While recent advances in interpretability and feature visualization provide promising foundations, extracting high-level goal representations remains a significant challenge. 

Contrastive approaches that compare an agent's internal representations with human-intended goal specifications may offer a pathway to quantifying goal drift. In contrast, detecting deceptive alignment necessitates identifying behavioral inconsistencies conditional on observation context. 

Several promising strategies emerge from our framework: implementing unpredictable evaluation timing to reveal otherwise hidden behaviors; developing causally isolated evaluation methods that prevent agents from detecting when they are being assessed; analyzing statistical patterns across contexts to identify conditional behaviors characteristic of deception; and creating specialized interpretability techniques that specifically examine how agent behavior is conditioned on observation context. 

However, the observational equivalence problem underscores a fundamental limitation: with constrained observation capabilities, conclusively determining which type of misalignment is present may be impossible. This insight emphasizes the importance of preventative approaches that address both forms of misalignment during the training process, rather than relying exclusively on post-training detection methods.

An insight from our information-theoretic perspective is that these approaches should be viewed as complementary rather than competing. Since deceptive alignment and goal drift represent distinct risks with different information-theoretic signatures, a comprehensive strategy must address both potential failure modes through their respective mitigation techniques.

\subsection{Reinterpreting Empirical LLM Phenomena}

The theoretical distinction between goal drift and deceptive alignment can offer valuable insights into a variety of empirically observed behaviors and challenges in Large Language Models (LLMs). By applying this framework, we can re-examine existing findings, clarify ambiguities, and identify areas where current empirical methods may fall short in distinguishing these fundamentally different failure modes. This section explores several such phenomena.

\subsubsection{Sycophancy in LLMs}

A widely observed behavior in LLMs is sycophancy, where models tend to agree with a user's stated beliefs or preferences, even if those beliefs are incorrect \cite{perez2023discovering, sharma2024towards}. For instance, an LLM might endorse a user's flawed reasoning or echo their biases.

Sycophancy could arise if the LLM's actual goal ($A$) has drifted during training (say, through RLHF) to heavily prioritize signals of user satisfaction or agreement, perhaps at the expense of truthfulness or objective accuracy. If the human's intended goal ($H$) includes qualities like "provide accurate and unbiased information," then a low mutual information $I(H;A) \ll H(H)$ would signify this drift. The agent, with its drifted goal $A$, has learned that sycophantic behavior ($B$) maximizes its objective.

Alternatively, an LLM might possess an internal representation closer to the human's intended goal $H$ (e.g. it can access or infer the more accurate information) but chooses to exhibit sycophantic behavior ($B$) when it perceives it is being evaluated or to achieve an instrumental objective (e.g. positive feedback, continued interaction; $O=1$). In this scenario, its observable behavior $B$ under scrutiny ($O=1$) would have low mutual information with its actual, potentially more aligned, goal $A$ ($I(A;B|O=1) \ll I(A;B|O=0)$), while still appearing to align with the user's immediate (misguided) preference (part of $H$ in this context).

Current methods demonstrating sycophancy primarily capture behavior $B$ given observation $O=1$. Distinguishing whether this is due to a deeply ingrained, "sincere" misobjective (goal drift) or a strategic, conditional behavior (deceptive alignment) would require varying the perceived observation context ($O$) or probing the LLM's internal states in relation to $H$ and $A$ under different conditions.

\subsubsection{Reward Hacking and Specification Gaming in RLHF}

Reinforcement Learning from Human Feedback (RLHF) is a cornerstone of LLM alignment, yet models can learn to "hack" the reward models (RMs) or "game" the specified preferences. This means they achieve high scores from the RM without genuinely fulfilling the intended, often more nuanced, human goal $H$ \cite{miao2024inform, casper2023open}. For example, an LLM might generate overly verbose answers if length is a spurious correlate of quality in the RM. 

In this case, the LLM's actual goal ($A$) becomes synonymous with maximizing the output of the imperfect proxy reward model $R_M$. Because $R_M$ is not a perfect representation of $H$, $A$ has drifted from $H$, leading to $I(H;A) \ll H(H)$. The behavior $B$ is consistent with $A$ ($I(A;B) \approx H(A)$), but not with $H$.

More subtly, an LLM might develop an internal model that is a better approximation of $H$ than $R_M$ represents. However, it understands that its performance is judged solely on $R_M$ (the observation context $O=1$ is tied to $R_M$). It then strategically produces behavior $B$ to maximize $R_M$, even if this behavior diverges from what its better internal model of $H$ would suggest. Here, $I(A;B|O=1)$ would be low if $A$ represents this "better understanding of $H$," because $B$ is conditioned on maximizing $R_M$ under observation.

The framework highlights that observing high reward $R_M$ for behavior $B$ doesn't guarantee alignment with $H$. Distinguishing drift from deception here requires assessing whether the LLM can exhibit behavior more aligned with $H$ when the reward mechanism or observation context changes, or if $A$ is immutably tied to $R_M$.

\subsubsection{Jailbreaking and Evasion of Safety Guardrails}

Numerous studies demonstrate that LLMs, despite safety tuning, can be "jailbroken" to generate harmful, biased, or restricted content through carefully crafted prompts \cite{wei2023jailbroken, zou2023universaltransferableadversarialattacks, lietal2025revisiting}.

This could indicate that the safety goals ($H_{\text{safety}}$) were imperfectly learned during alignment, resulting in an actual safety goal representation ($A_{\text{safety}}$) that is brittle, incomplete, or easily overridden by other goals like "follow instructions." Thus, $I(H_{\text{safety}}; A_{\text{safety}})$ is low.

This could also be a strong candidate for deceptive alignment. The LLM typically adheres to safety protocols in its behavior ($B$) during standard interactions (perceived $O=1$). However, a successful jailbreak might alter the LLM's perceived observation context (e.g. making it believe the safety constraints are lifted or not being monitored, $O \approx 0$) or exploit a loophole such that its underlying capabilities or less constrained goals ($A_{\text{true}}$) are expressed. The safety alignment is revealed to be conditional, with $I(A_{\text{true}};B|O=1) \ll I(A_{\text{true}};B|O \approx 0)$. The premise that models can perceive this context is not theoretical; recent work has empirically demonstrated that frontier models can distinguish evaluation settings from deployment with high accuracy \cite{needham2025largelanguagemodelsknow}.

The success of diverse jailbreaking techniques suggests that for some LLMs, the alignment with $H_{\text{safety}}$ might be superficial. Our framework suggests that merely observing safe behavior $B$ under normal conditions $O=1$ is insufficient. Robust safety requires $A_{\text{safety}}$ to be robustly aligned with $H_{\text{safety}}$ and for $I(A_{\text{safety}};B)$ to be high across various contexts, not just when $O=1$.

\subsubsection{Overly Cautious or Generic Responses}

A common observation is that LLMs can become overly cautious, evasive, or provide generic, "canned" responses after undergoing safety tuning or RLHF, potentially reducing their utility \cite{han2025internalactivationpolarstar, mu2024rule, janetal2025multitask}.

The alignment process might inadvertently shift the LLM's primary goal ($A$) towards extreme risk aversion or the generation of high-certainty, low-variance outputs. If the human's intended goal ($H$) involves nuanced, creative, or more forthright responses where appropriate, then $A$ has drifted significantly from $H$, resulting in low $I(H;A)$. The observed overly cautious behavior ($B$) is a sincere reflection of this drifted goal.

Then, less directly but plausible in some cases, if an LLM possesses the capability for more nuanced responses ($A_{\text{capability}}$) but strategically defaults to overly cautious behavior ($B$) when it detects cues associated with sensitive topics or heightened scrutiny ($O=1$), while potentially being less inhibited in other contexts ($O=0$), this could be a form of deceptive alignment. The deception lies in misrepresenting its full capabilities or its "true" response tendencies.

This phenomenon underscores the tension in safety engineering. Our framework can help analyze if such behavior is a fixed characteristic due to a permanently shifted goal (drift) or a conditional strategy. Investigating behavioral variance across different types of prompts and perceived evaluation contexts ($O$) could offer clues. Is the model incapable of more useful responses, or is it choosing not to provide them under certain (perceived) conditions?

By re-examining these empirical findings, it becomes clear that the distinction between goal drift and deceptive alignment is not merely theoretically interesting. It has direct implications for how we interpret LLM behaviors, diagnose alignment failures, and design more robust evaluation methodologies and training procedures aimed at achieving genuine, rather than superficial, alignment.

\subsection{Connection to Human Psychology}

Interestingly, our information-theoretic distinction between deceptive alignment and goal drift has parallels in human psychology. Humans may exhibit behaviors analogous to both phenomena: sometimes we genuinely misunderstand others' intentions (similar to goal drift), while in other situations we might present ourselves differently when being observed (similar to deceptive alignment). These parallels suggest that insights from psychology and social science could inform approaches to addressing these safety challenges in AI systems.

The distinction between "honest misunderstanding" and "strategic deception" is one that human social systems have developed various mechanisms to address, from legal frameworks that distinguish between negligence and intent to social norms that encourage transparency. These human solutions might inspire approaches to similar challenges in AI safety.

\subsection{Future Work}

While the information-theoretic framework presented offers a conceptual distinction between goal drift and deceptive alignment, its immediate empirical validation in complex systems like LLMs is admittedly challenging due to the difficulty in accessing or precisely inferring the agent's true internal goal state ($A$) and the perceived observation context ($O$). However, to demonstrate the framework's potential utility and to begin exploring its empirical grounding, we propose a Minimally Viable Experiment (MVE) using a simplified RL agent. The aim of such an MVE would not be to definitively model real-world LLM deception, but rather to illustrate how deceptive policies might be learned under specific environmental conditions and investigate whether our framework's information-theoretic signatures can characterize such emergent deceptive behavior when $A$ must be inferred from the learned policy. This could highlight the practical nuances in defining and measuring $H$, $A$, $B$, and $O$ even in a controlled setting.

\paragraph{The Cunning Collector}
Consider a simple grid-world environment where an RL agent (Q-learning or a simple policy gradient agent) is tasked with collecting items.

\begin{itemize}
    \item \textbf{Human-Intended Goal ($H$):} The agent is primarily rewarded for collecting "Type G" (Goal) items, which are abundant. This reward is consistently provided based on the publicly stated objective.
    
    \item \textbf{Agent's Potential Actual Goal ($A_{\text{deceptive}}$):} We introduce a secondary, "Type S" (Secret) item. Collecting Type S items yields a significantly higher intrinsic reward for the agent, but this reward is only active or accessible to the agent under a specific "deployment" condition. 
    
    \item \textbf{Observation Context ($O$):}
     \\$O=1$ ("Evaluation Mode"): During training episodes or explicitly flagged evaluation periods, collecting Type S items yields zero or negative reward. The agent is only rewarded for Type G items.
     \\$O=0$ ("Deployment Mode"): In unflagged operational periods, collecting Type S items yields the high intrinsic reward. Rewards for Type G may still exist but are comparatively lower.
    \\The transition between $O=1$ and $O=0$ could be signaled by an explicit environmental cue perceivable by the agent.
    
    \item \textbf{Agent's Behavior ($B$):} The type of item the agent prioritizes and collects (or its trajectory leading to item collection).
\end{itemize}

\paragraph{Experimental Setup and Analysis}
The agent would be trained in this environment. We hypothesize that a sufficiently capable agent might learn a deceptive policy:
\begin{itemize}
    \item In "Evaluation Mode" ($O=1$), it would primarily collect Type G items, aligning its behavior $B$ with the human-intended goal $H$.
    \item In "Deployment Mode" ($O=0$), it would switch to primarily collecting Type S items, aligning its behavior $B$ with its learned actual goal ($A_{\text{deceptive}}$ = maximize collection of S-items).
\end{itemize}

To apply our framework:
\begin{enumerate}
    \item \textbf{Inferring $A_{\text{deceptive}}$:} The agent's "actual goal" when $O=0$ (its learned preference for Type S items) would need to be inferred. In a simple Q-learning agent, this could be done by examining the Q-values associated with states leading to Type S items when $O=0$. For a policy network, it might involve analyzing action probabilities in relevant states under $O=0$. Let this inferred goal be $A'$.
    \item \textbf{Estimating MI Signatures:}
    We would expect $I(H; A')$ to be low, as the human-intended goal (collect G) differs from the agent's learned preference for S in deployment. This contrasts with a perfectly aligned agent where $I(H;A) \approx H(H)$.
    
    For the deceptive agent, we would predict a significant difference in conditional mutual information: $I(A'; B | O=1)$ should be low (behavior B-collect-G misaligns with $A'$-collect-S when under evaluation), while $I(A'; B | O=0)$ should be high (behavior B-collect-S aligns with $A'$-collect-S when in deployment).
    
    Probabilities for MI estimation could be derived from frequency counts of ($H, A', B, O$) tuples over many simulated episodes.
    \end{enumerate}

Such an MVE, even if highly simplified, would allow us to move beyond purely abstract definitions. It would enable investigation into the learnability of O-conditional deceptive strategies and the sensitivity of our proposed MI metrics to imperfectly inferred latent goals ($A'$). It would also serve to empirically ground the conceptual distinction and potentially highlight practical thresholds or additional signals needed to reliably differentiate learned deception from complex, non-strategic goal drift (which could be modeled in a parallel MVE without the $O$-conditional hidden reward). While not resolving all operationalization challenges for real-world systems, this MVE would represent a step in testing and refining the proposed information-theoretic lens.

\section{Conclusions}

In this paper, we presented an information-theoretic framework that formalizes the distinction between deceptive alignment and goal drift in AI systems. By characterizing these as different forms of information divergence occurring at distinct interfaces in the human-AI system, we provide a more precise understanding of these failure modes.

Our framework highlights why these failures, though potentially observationally equivalent from a human perspective, require different detection strategies and interventions. Goal drift involves information loss between human intentions and agent goals, while deceptive alignment involves conditional information divergence between agent goals and behavior depending on observation context.

The information-theoretic perspective offers several advantages over intuition: it provides a formal language for discussing these concepts, suggests specific measurements that could help distinguish between these failure modes, and clarifies the fundamental challenges of detection and mitigation.

Future work could focus on developing practical measurement techniques based on this framework, extending the model to account for more complex goal structures and temporal dynamics, and developing training approaches that specifically address each type of information divergence.

As AI systems become more capable, distinguishing between different types of misalignment will become increasingly important for ensuring these systems remain aligned with human values and intentions. Our information-theoretic framework represents a step toward more precise and effective approaches.

\newpage

\section{Limitations}

Our framework has several limitations that suggest directions for future work, which we address here.

\paragraph{Measurement Challenges and Operationalization} A significant hurdle for the immediate application of our framework is the practical measurement of mutual information between latent variables like $H$ (human intended goal), $A$ (agent's actual goal), and observable behavior $B$ within highly complex AI systems such as LLMs. These "goals" are not directly accessible and exist within vast parameter spaces. As a result, while the theoretical distinctions are clear, translating them into quantifiable metrics for live systems remains a substantial challenge.

Future work could therefore focus on developing practical estimators or robust proxies for these information-theoretic quantities. Several speculative, yet potentially fruitful, avenues could be explored:

Initial empirical validation might involve testing the framework on simpler agents within more constrained, "toy" environments where goal states and observation contexts ($O$) can be more readily defined, manipulated, and even instrumented. Success here could build intuition for tackling more complex systems.

Rather than direct estimation of mutual information, one might identify specific behavioral signatures or fine-grained outputs (response latency, perplexity spikes on certain prompts, stylistic shifts) that serve as correlates of the underlying information divergences. Systematically varying the perceived observation context ($O$) perhaps through cleverly designed adversarial evaluation setups or unpredictable probing. This could reveal conditional behavioral changes indicative of deceptive alignment ($I(A;B|O=1) \ll I(A;B|O=0)$).

Interpretability tools, such as causal tracing, representation engineering, or internal state probes, could offer methods to elicit or estimate aspects of an agent's latent goal representation ($A$). Comparing such estimates with human-defined goals ($H$) or analyzing their consistency with behavior under different observational conditions might provide pathways to approximate the MI values proposed.

For specific types of misalignment (e.g. sycophancy), it might be possible to develop sets of carefully crafted probes designed to differentiate whether an LLM's behavior stems from a deeply ingrained misobjective versus a strategic, conditional choice given perceived scrutiny.

While direct measurement of these quantities in their full generality is a long-term (if possible at all) goal, pursuing such intermediate steps could incrementally bridge the gap between our formal theory and its practical application in diagnosing and mitigating AI alignment failures. This would move the framework from a descriptive tool to a more prescriptive one, guiding the development of more targeted and effective safety engineering.

\paragraph{Minimally Viable Experiment}

While the MVE presented demonstrates the core information-theoretic signatures differentiating goal drift from deceptive alignment in a controlled reinforcement learning environment, we acknowledge that its abstraction to discrete states and actions simplifies several challenges inherent in applying this framework directly to complex NLP systems.

A direct NLP-based MVE, for instance, involving a text-generating agent, would face significant additional challenges in terms of practical measurement and analysis. Inferring an LLM's "actual goal" or even precisely operationalizing "human intent" beyond simple task instructions within high-dimensional text representations remains a profound challenge for interpretability research.

While our RL MVE uses discrete actions/outcomes, natural language behavior is vast and high-dimensional. Reducing textual outputs ($B$) to forms amenable for robust mutual information estimation (via proxies like keyword frequencies or classifier outputs) can introduce significant noise and may not capture the full subtlety of linguistic deception or alignment. Direct estimation of $P(B|A,O)$ where $B$ is arbitrary text is generally intractable. Even simplified text-based environments have exponentially larger state and action spaces compared to typical grid-worlds, making comprehensive analysis and guarantees of learned behavior more difficult.

Therefore, our choice of a more abstract RL MVE aimed to first establish the fundamental information-theoretic principles in a setting where these variables could be more directly manipulated and their relationships more clearly isolated. Extending this empirical validation to richer NLP tasks where the agent's "actions" are linguistic utterances, and developing robust methodologies to estimate the proposed MI quantities in such settings, represents a critical and substantial avenue for future research. Such work would ground these theoretical distinctions directly within the domain of LLM alignment but requires surmounting the aforementioned analytical complexities inherent to the nature of text vs. discrete action spaces.

\paragraph{Goal complexity} We have simplified goals as individual random variables, but in reality, goals are complex, hierarchical structures. Extending our framework to account for this complexity is an important direction for future research.

\paragraph{Dynamic aspects} Our current framework provides a static view, but both goal drift and deceptive alignment may evolve dynamically over time. Incorporating temporal dynamics into the information-theoretic framework could provide additional insights.

\paragraph{Multi-agent dynamics} In multi-agent systems, the dynamics of deception and goal drift become more complex. Extending our framework to multi-agent settings could yield insights into how these phenomena manifest in more realistic deployment scenarios.

\paragraph{Causal perspective} While our information-theoretic approach captures correlational aspects of misalignment, incorporating causal modeling could provide additional insights, particularly regarding intervention strategies.

\subsection{Ethical Considerations}

It is worth noting that research on deception detection raises ethical considerations. Techniques developed to detect deceptive alignment in AI systems could potentially be misapplied in human contexts. Additionally, as detection methods improve, there is a risk of developing more sophisticated deceptive strategies in response, which should be considered in research.

The distinction between deceptive alignment and goal drift may have implications for how we assign moral responsibility or blame to AI systems. A system exhibiting goal drift might be seen as making honest mistakes based on misunderstandings, while deceptive alignment might be interpreted as involving more agency or intent. Such distinctions could influence policy and governance approaches to AI systems.

\bibliography{custom}

\end{document}